\documentclass[runningheads]{llncs}
\usepackage{graphicx}
\usepackage{tikz}
\usepackage{comment}
\usepackage{amsmath,amssymb} 

\usepackage{threeparttable}
\usepackage{pifont}
\usepackage{booktabs}
\usepackage{wrapfig}
\newsavebox\CBox
\def\textBF#1{\sbox\CBox{#1}\resizebox{\wd\CBox}{\ht\CBox}{\textbf{#1}}}

\usepackage{ulem}

\usepackage[accsupp]{axessibility}  

\begin{document}
\pagestyle{headings}
\mainmatter

\title{SocialVAE: Human Trajectory Prediction\\
using Timewise Latents}


\titlerunning{SocialVAE: Human Trajectory Prediction using Timewise Latents}
%
\author{
Pei Xu\inst{1,3} \and
Jean-Bernard Hayet\inst{2} \and
Ioannis Karamouzas\inst{1}
}%
\authorrunning{P. Xu et al.}
%
\institute{
Clemson University, South Carolina, USA \and
CIMAT, A.C., M\'{e}xico \and 
Roblox \\
\email{peix@clemson.edu},
\email{jbhayet@cimat.mx},
\email{ioannis@clemson.edu} \\
\texttt{https://motion-lab.github.io/SocialVAE}
}
\maketitle

\begin{abstract}
Predicting pedestrian movement is critical for human behavior analysis and also for safe and efficient human-agent interactions. 
However, despite significant advancements, it is still challenging for existing approaches to capture the uncertainty and multimodality of human navigation decision making. 
In this paper, we propose SocialVAE, a novel approach for human 
trajectory prediction. The core of SocialVAE is a timewise variational autoencoder architecture that exploits stochastic recurrent neural networks to perform prediction,  
combined with a social attention mechanism and a backward posterior approximation to allow for better extraction of pedestrian navigation strategies.
We show that SocialVAE improves current state-of-the-art performance on several pedestrian trajectory prediction benchmarks,
including the ETH/UCY benchmark, Stanford Drone Dataset, and SportVU NBA movement dataset.
\end{abstract}

\section{Introduction}
The development of autonomous agents interacting with humans,
such as self-driving vehicles, indoor service robots, and 
traffic control systems,
involves human  behavior modeling and inference in order to meet both the safety and intelligence requirements.
In recent years, with the rise of deep learning techniques,
extracting patterns from sequential data for prediction or sequence transduction has advanced significantly in fields such as machine translation~\cite{vaswani2017attention,brown2020language}, image completion~\cite{van2016pixel,van2016conditional}, weather forecasting~\cite{ravuri2021skilful,weyn2019can}, and physical simulation~\cite{sanchez2020learning,kochkov2021machine}.
In contrast to works performing inference from regularly distributed data conforming to specific rules or physics laws,
predicting human behaviors, e.g., body poses and socially-aware movement, still faces huge challenges due to the complexity and uncertainty in the human decision making process.

In this work, we focus on the task of human trajectory prediction from short-term historical observations.
Traditional works predict pedestrian trajectories using deterministic models~\cite{helbing1995social,helbing,pellegrini2009you,prl,orca}.
However, human navigation behaviors have an inherent multimodal nature with lots of randomness.
Even in the same scenario, there would be more than one trajectories that a pedestrian could take.
Such uncertainty cannot be captured effectively by deterministic models,
especially for long-term trajectory prediction with more aleatory influences introduced.
Furthermore, individuals often exhibit different behaviors
in similar scenarios.
Such individual differences are decided by various stationary and dynamical factors such as crowd density and scene lighting, weather conditions, social context, personality traits, etc.  
As such, the complexity of behaviors is hard to be consistently modeled by rule-based methods, which work under predetermined physical laws and/or social rules~\cite{helbing1995social,helbing,prl,pradhan2011robot}.
Recent works have promoted data-driven solutions based on deep generative models to perform stochastic predictions or learn the trajectory distribution directly~\cite{sgan2018,sattention2018,sadeghian2019sophie,pecnet2020,ivanovic2019trajectron,salzmann2020trajectron++,wang2022stepwise,yao2021bitrap,yuan2021agentformer}.
Despite impressive results, current approaches still face the challenge of making high-fidelity predictions with a limited number of samples.

In this paper, we exploit recent advances in variational inference techniques and introduce a timewise variational autoencoder (VAE) architecture for human trajectory prediction. 
Similar to prior VAE-based methods~\cite{pecnet2020,salzmann2020trajectron++,wang2022stepwise,yao2021bitrap,yuan2021agentformer}, we rely on recurrent neural networks (RNNs) to handle trajectory data sequentially and provide stochastic predictions.
However, 
our model uses latent variables as stochastic parameters 
to condition the hidden dynamics of RNNs at \textit{each time step}, in contrast to previous solutions that condition the prior of latent variables only based on historical observations. 
This allows us to more accurately capture the dynamic nature of human decision making.
Further, to robustly extract navigation patterns from whole trajectories, we use a backward RNN structure for posterior approximation. 
To cope with an arbitrary number of neighbors during observation encoding, we develop a human-inspired attention mechanism to encode neighbors' states by considering the observed social features exhibited by these neighbors. 

Overall, 
this paper makes the following contributions:
\begin{itemize}
    \setlength\itemsep{0.25em}
    \item We
    propose SocialVAE, a novel approach to predict human trajectory distributions conditioned on short-term historical observations. 
    Our model employs a \textit{timewise} VAE architecture with a conditional prior and a posterior approximated bidirectionally from the whole trajectory, and uses an attention mechanism to capture the social influence from the neighboring agents. 
    \item We introduce \textit{Final Position Clustering} as an optional and easy-to-implement postprocessing technique to help reduce sampling bias and further improve the overall prediction quality when a limited number of samples are drawn from the predicted distribution.
    \item We experimentally show that SocialVAE captures the mutimodality of human navigation behaviors while reasoning about human-human interactions in both everyday settings and NBA scenarios involving cooperative and adversarial agents with fast dynamics.
    \item 
     We achieve state-of-the-art performance on the ETH/UCY and SDD benchmarks and SportVU NBA movement dataset, bringing more than $10\%$ improvement and in certain test cases more than $50\%$ improvement over existing trajectory prediction methods.
\end{itemize}

\section{Related Work}
Research in pedestrian trajectory prediction can be broadly classified into human-space and human-human models.
The former focuses on predicting scene-specific human movement patterns~\cite{kitani2012activity,ballan2016knowledge,sadeghian2018car,cao2020long,mangalam2021goals} and takes advantage of the scene environment information, typically through the use of semantic maps.
In this work, we are interested in the latter, which performs trajectory prediction by using dynamic information about human-human interactions.

\noindent\textbf{Mathematical Models.}
Modeling human movement in human-human interaction settings 
typically leverages hand-tuned mathematical models to perform deterministic prediction. 
Such models include 
rule-based approaches using social forces, velocity-obstacles, and energy-based formulations~\cite{helbing1995social,orca,yamaguchi2011you,prl}. 
Statistical models based on observed data such as Gaussian processes~\cite{wang2007gaussian,trautman2010unfreezing,kim2011gaussian} have also been widely used. By nature, they cope better with uncertainty on overall trajectories but struggle on fine-grained variations due to social interactions. 

\noindent\textbf{Learning-based Models.}
In recent years, data-driven methods using deep learning techniques
for trajectory prediction have achieved impressive results.
SocialLSTM~\cite{slstm2016} employs a vanilla RNN structure using long short-term memory (LSTM) units to perform prediction sequentially.
SocialAttention~\cite{sattention2018} introduces an attention mechanism to capture neighbors' influence by matching the RNN hidden state of each agent to those of its neighbors.
SocialGAN~\cite{sgan2018}, SoPhie~\cite{sadeghian2019sophie} and SocialWays~\cite{sways2019} use generative adversarial network (GAN) architectures. 
To account for social interactions,  
SocialGAN proposes a pooling process to synthesize neighbors via their RNN hidden states, 
while SocialWays adopts an attention mechanism that takes into account the neighbors' social features.
PECNet~\cite{pecnet2020}, Trajectron~\cite{ivanovic2019trajectron}, Trajectron++~\cite{salzmann2020trajectron++},
AgentFormer~\cite{yuan2021agentformer},
BiTraP~\cite{yao2021bitrap} and SGNet~\cite{wang2022stepwise} employ a conditional-VAE architecture~\cite{sohn2015learning} to predict trajectory distributions,
where latent variables are generated conditionally to
the given observations.
Memory-based approaches for trajectory prediction have also been recently explored such as MANTRA~\cite{marchetti2020mantra} and MemoNet~\cite{xu2022remember}. 
Though achieving impressive results, such approaches would typically suffer from 
slow inference speeds and storage issues when dealing with large scenes. 
Recent works~\cite{giuliari2021transformer,yu2020spatio} have also exploited Transformer architectures to perform trajectory prediction. However, as we will show in Section~\ref{sec:exp}, Transformer-based approaches tend to have worse performance than VAE-based approaches. 

\noindent\textbf{Stochastic RNNs. }
To better model highly dynamic and multimodal data,
a number of recent works~\cite{storn2014,vrnn2015,srnn2016,zforcing2017} leverage VAE architectures to extend RNNs with timewisely generated 
stochastic latent variables.
Despite impressive performance on general sequential data modeling tasks, 
these approaches do not consider the interaction features appearing in human navigation tasks. 

Following the literature of stochastic RNNs, we propose to use a timewise VAE as the backbone architecture for human trajectory prediction.
The main motivation behind our formulation is that human decision making is highly dynamic and can lead to different trajectories at any given time. 
Additionally, to better extract features in human-human interactions, 
we employ a backward RNN for posterior approximation, which takes the \textit{whole} ground-truth (GT) trajectory into account during learning. 
Neighbors are encoded through an attention mechanism that uses social features similar to~\cite{sways2019}.
A major advantage of our attention mechanism is that it relies only on the neighbors' observable states (position and velocity). In contrast, previous attention-based works use RNN hidden states as the representation of neighbors' states~\cite{sgan2018,sways2019,sadeghian2019sophie},
which can only take into account neighbors that are consistently tracked during observation.
As we show, our timewise VAE with the proposed attention mechanism achieves state-of-the-art performance on ETH/UCY/NBA datasets.

\section{Approach}
Our proposed SocialVAE approach infers the distribution of future trajectories for each agent in a scene 
based on given historical observations.
Specifically, given a scene containing $N$ agents, let $\mathbf{x}_i^t \in \mathbb{R}^2$ be the 2D spatial coordinate of agent $i$ at time step $t$. 
We perform $H$-frame inference for the distribution over the agent's future positions  $\mathbf{x}_i^{\scriptscriptstyle T+1:T+H}$ based on a $T$-frame joint observation, 
i.e., we estimate $p(\mathbf{x}_i^{\scriptscriptstyle T+1:T+H}\vert\mathcal{O}_{i}^{\scriptscriptstyle 1:T})$ where $\mathcal{O}_{i}^{\scriptscriptstyle 1:T}$
gathers the local observations from agent $i$ to the whole scene.
SocialVAE performs prediction for each agent independently, based on social features extracted from local observations, and can run with scenes having an arbitrary number of agents.  
This may be of particular interest in real-time and highly dynamic environments where the local neighborhood of a target agent is constantly changing or cannot be tracked consistently. 

\subsection{Model Architecture}
\label{subsec:architecture}
\begin{figure*}[t]
    \centering
    \includegraphics[width=\linewidth]{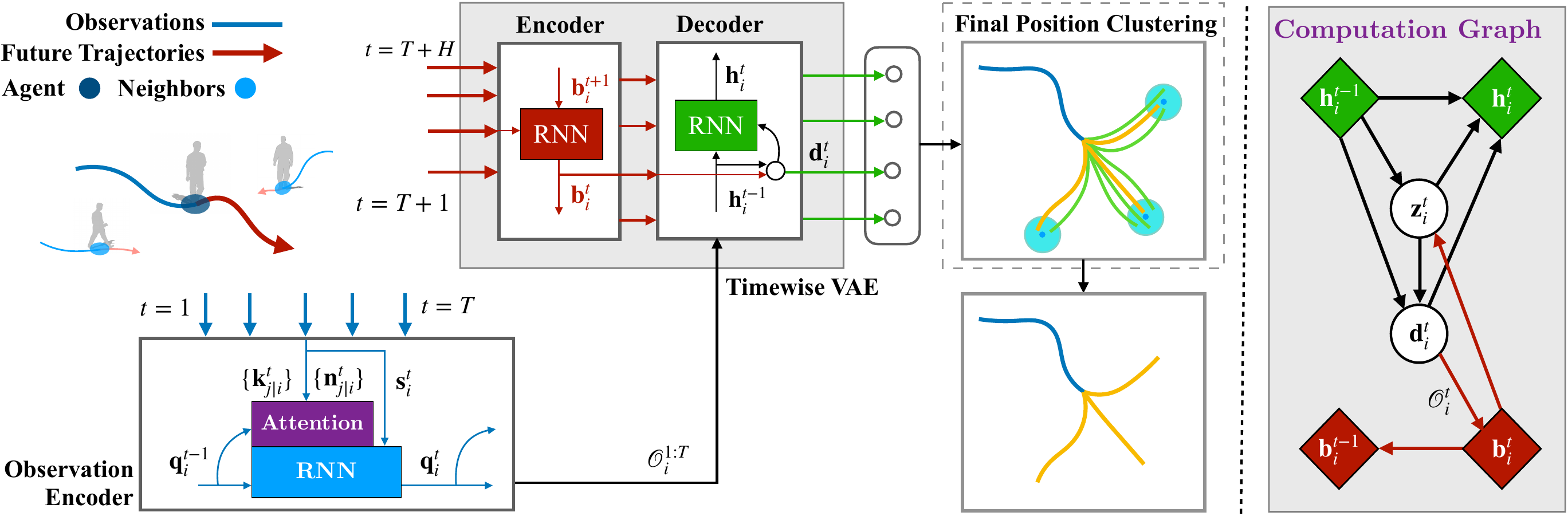}
    \caption{
    Overview of SocialVAE that uses
    an 
    RNN-based timewise VAE with sequentially generated stochastic latent variables for trajectory prediction. SocialVAE can be coupled with a \textit{Final Position Clustering} postprocessing scheme
    to improve the 
    prediction 
    quality. 
    The observation encoder attention mechanism considers each neighbor's state $\mathbf{n}_{j \vert i}$ along with its social features $\mathbf{k}_{j \vert i}$.
    The computation graph (right) shows the state transfer inside the timewise VAE. Diamonds represent deterministic states and circles represent stochastic states. Red parts are used only at training. 
    }
    \label{fig:overview}
\end{figure*}
Figure~\ref{fig:overview} shows the system overview of our model. 
Its backbone is a timewise VAE.
As in prior works~\cite{zforcing2017,storn2014,vrnn2015,srnn2016}, 
we use 
an 
RNN structure to condition the sequential predictions through an auto-regressive model relying on the state variable of the RNN structure. 
However, while prior works directly perform predictions over time-sequence data,
we introduce the past observations as \textit{conditional variables}.
Moreover, instead of directly predicting the absolute coordinates $\mathbf{x}_i^{T+1:T+H}$, we generate a displacement sequence $\mathbf{d}_i^{T+1:T+H}$, 
where $\mathbf{d}_i^{t+1} \triangleq \mathbf{x}_i^{t+1} - \mathbf{x}_i^t$.
The target probability distribution of the displacement sequence can be written as
\begin{equation}\label{eq:v_dist}
     p(\mathbf{d}_i^{T+1:T+H}\vert\mathcal{O}_{i}^{1:T}) = \prod_{\tau=1}^{H} p(\mathbf{d}_i^{T+\tau}\vert\mathbf{d}_{i}^{T+1:T+\tau-1},\,\mathcal{O}_{ i}^{1:T}).
\end{equation}
To generate stochastic predictions, we use a conditional prior over the RNN state variable to introduce latent variables at each time step,
and thus obtain a timewise VAE architecture, allowing us to model highly nonlinear dynamics during multi-agent navigation.
In the following, we describe our generative model for trajectory prediction and the inference process for posterior approximation. 

\noindent{\textbf{Generative Model. }}
Let $\mathbf{z}_i^t$ be the latent variables introduced at  time step $t$.
To implement the sequential generative model $p(\mathbf{d}_i^t|\mathbf{d}_i^{t-1},\mathcal O_i^{1:T},\mathbf{z}_i^{t})$,
we use
an 
RNN in which the state variable $\mathbf{h}_i^t$ is updated recurrently by
\begin{equation}
    \mathbf{h}_i^t = \overrightarrow{g}(\psi_\mathbf{zd}(\mathbf{z}_i^t, \mathbf{d}_i^t), \mathbf{h}_i^{t-1}), 
\end{equation} 
for $t=T+1,\ldots,T+H$. 
In the recurrence, the initial state is extracted from historical observations, i.e.,  
$\mathbf{h}_i^T = \psi_{\mathbf{h}}(\mathcal{O}_{i}^{1:T})$,  
where 
$\psi_\mathbf{zd}$ and $\psi_{\mathbf{h}}$ are two
embedding neural networks. 
Developing Eq.~\ref{eq:v_dist} with $\mathbf{z}_i^t$
, we obtain the generative model:
\begin{equation}\small
    p(\mathbf{d}_i^{T+1:T+H}\vert\mathcal{O}_{i}^{1:T}) 
    = \prod_{t=T+1}^{T+H} \int_{\mathbf{z}_i^{t}} 
    p(\mathbf{d}_i^{t}\vert\mathbf{d}_{i}^{T:t-1},\,\mathcal{O}_{ i}^{1:T},\mathbf{z}_i^{t})
    p(\mathbf{z}_i^{t}\vert\mathbf{d}_{i}^{T:t-1},\,\mathcal{O}_{ i}^{1:T})   \text{d}\mathbf{z}_i^{t}.
\label{eq:joint}
\end{equation}

In contrast to standard VAEs that use a standard normal distribution as the prior,
our prior distribution is conditioned and can be obtained from the RNN state variable. 
The second term of the integral in Eq.~\ref{eq:joint} can be translated into
\begin{equation}
    p(\mathbf{z}_i^{t}\vert\mathbf{d}_{i}^{T:t-1},\,\mathcal{O}_{ i}^{1:T}) := p_{\theta}(\mathbf{z}_i^{t}\vert\mathbf{h}_i^{t-1}),
    \label{eq:timewisely}
\end{equation}
where $\theta$ are parameters for a neural network that should be optimized.
This results in a parameterized conditional prior distribution over the RNN state variable, through which we can track the distribution flexibly. 

The first term of the integral in Eq.~\ref{eq:joint} implies sampling new displacements from the prior distribution $p$,
which is conditioned on the latent variable $\mathbf{z}_i^{t}$, and on the observations and previous displacements captured by $\mathbf{h}_i^{t-1}$, i.e. 
\begin{equation}
    \mathbf{d}_i^t \sim p_\xi(\cdot\vert\mathbf{z}_i^t,\mathbf{h}_i^{t-1}),
\end{equation}
with parameters $\xi$.
Given the displacement definition $\mathbf{d}_i^t$, we obtain  
\begin{equation}\label{eq:x_recurrent}
    \mathbf{x}_i^t = \mathbf{x}_i^{T} + \sum\nolimits_{\tau=T+1}^{t} \mathbf{d}_i^\tau, 
\end{equation}
as a stochastic prediction for the spatial position of agent $i$ at time $t$.

\noindent{\textbf{Inference Model. }} 
To approximate the posterior $q$ over the latent variables, we 
consider the whole GT observation
$\mathcal{O}_{i}^{1:T+H}$
to shape the latent variable distribution via a backward recurrent network~\cite{srnn2016,zforcing2017}:
\begin{equation}
    \mathbf{b}_i^t = \overleftarrow{g}(\mathcal{O}_{i}^t,\mathbf{b}_i^{t+1}),
\label{eq:backwards}    
\end{equation}
for $t=T+1,\cdots,T+H$ given the initial state $\mathbf{b}_i^{T+H+1} = \mathbf{0}$.
The state variable $\mathbf{b}_i^t$ provides the GT trajectory information backward from $T+H$ to $t$.
By defining the posterior distribution as a function of both the backward state $\mathbf{b}_{i}^t$ and forward state $\mathbf{h}_{i}^t$, 
the latent variable $\mathbf{z}_{i}^t$ is
drawn implicitly during inference based on the entire GT trajectory. 
With $\phi$ denoting  the parameters of the network mapping $\mathbf{b}_{i}^t$ and $\mathbf{h}_{i}^{t-1}$ to the posterior parameters, we can sample latent variables as
\begin{equation}
    \mathbf{z}_{i}^t \sim q_\phi(\cdot \vert \mathbf{b}_{i}^t, \mathbf{h}_{i}^{t-1}).
\end{equation}

The computation graph shown in Fig.~\ref{fig:overview} gives an illustration of the dependencies of our generative and inference models.
Note that the inference model (red parts in Fig.~\ref{fig:overview}) is employed only at training. During testing or evaluation, only the generative model coupled with the observation encoding module is used to perform predictions, and no information from future trajectories is considered.

\noindent{\textbf{Training. }}
Similarly to the standard VAE, the learning objective of our model is to maximize the evidence lower bound (ELBO) that sums up over all time steps given the target distribution defined in Eq.~\ref{eq:joint}:
\begin{equation}\label{eq:elbo_v}\small
    \sum_{t=T+1}^{T+H}\mathbb{E}_{\mathbf{z}_{i}^t \sim q_\phi(\cdot \vert \mathbf{b}_{i}^t, \mathbf{h}_{i}^{t-1})}\left[ \log p_\xi(\mathbf{d}_i^t \vert \mathbf{z}_i^t,\mathbf{h}_i^{t-1}) \right] 
    - D_{KL}\left[q_\phi(\mathbf{z}_i^t \vert \mathbf{b}_i^t, \mathbf{h}_i^{t-1}) \vert\vert p_\theta(\mathbf{z}_i^t \vert \mathbf{h}_i^{t-1})\right].
\end{equation}
where $p_\xi$, $q_\phi$ and $p_\theta$ are parameterized as Gaussian distributions by networks.

Optimizing Eq.~\ref{eq:elbo_v} with the GT value of $\mathbf{d}_i^t$ ignores accumulated errors when we project $\mathbf{d}_i^t$ back to $\mathbf{x}_i^t$ 
to get the final trajectory (Eq.~\ref{eq:x_recurrent}).
Hence, we replace the log-likelihood term with the squared error over $\mathbf{x}_i^t$ and optimize over $\mathbf{d}_i^t$ through reparameterization tricks~\cite{Kingma14}, for prediction errors in previous time steps to be compensated in next time steps.
The final training loss is 
\begin{equation}\small
    \mathbb{E}_{i}\Biggl[ \frac{1}{H} \sum_{t=T+1}^{T+H} 
    \mathbb{E}_{\substack{
    \mathbf{d}_i^t \sim p_\xi(\cdot\vert\mathbf{z}_i^t,\mathbf{h}_i^{t-1})\\
    \mathbf{z}_i^t \sim q_\phi(\cdot\vert\mathbf{b}_i^t,\mathbf{h}_i^{t-1})}
    }\biggl[ \| (\mathbf{x}_i^t - \mathbf{x}_i^T) - \sum_{\substack{\tau=\\T+1}}^t \mathbf{d}_i^\tau \|^2 \\
    \text{} + q_\phi(\mathbf{z}_i^t\vert\mathbf{b}_i^t,\mathbf{h}_i^{t-1})
    - p_\theta(\mathbf{z}_i^t\vert\mathbf{h}_i^{t-1})
    \biggr]\Biggr].
\end{equation}
For simplicity, at training, we sample $\mathbf{z}_i^t$ and $\mathbf{d}_i^t$ only once every time,
and use the reparameterization trick of Gaussian distributions to update $q_\phi$, $p_\xi$ and $p_\theta$.

\subsection{Observation Encoding}\label{subsec:observation_encoding} 
Consider a scene with an agent $i$ being the target of our prediction process and multiple neighboring agents (their number may vary along the observation sequence of agent $i$).  
We define the local observation from agent $i$ to the whole scene at time step $t = 2, \cdots, T$ as the vector containing the observation to the agent itself and the synthesis of all its neighbors: 
\begin{equation}
    \mathcal{O}_{i}^t := \left[ f_\mathbf{s}(\mathbf{s}_i^t), \sum\nolimits_j w_{j \vert i}^t f_\mathbf{n}(\mathbf{n}_{j \vert i}^t) \right],
\end{equation}
where
\begin{itemize}
    \item $\mathbf{s}_i^t := \left[\mathbf{d}_i^t, \mathbf{d}_i^t - \mathbf{d}_i^{t-1} \right] \in \mathbb{R}^4$ is the self-state of agent $i$, including the agent's velocity and acceleration information represented by position displacement,
    \item $\mathbf{n}_{j \vert i}^t := \left[ \mathbf{x}_j^t-\mathbf{x}_i^t, \mathbf{d}_j^t-\mathbf{d}_i^t \right] \in \mathbb{R}^4$ is the local state of neighbor agent $j$ relative to agent $i$, including its relative position and velocity,
    \item $f_\mathbf{s}$ and $f_\mathbf{n}$ are learnable feature extraction neural networks,
    \item $w_{j \vert i}^t$ is an attention weight through which features from an arbitrary number of neighbors are fused together into a fixed-length vector. 
\end{itemize}
Neighbors are re-defined at every time step: agent $j$ is a neighbor of agent $i$ at time step $t$ if $j \neq i$ and $\vert\vert \mathbf{x}_j^t - \mathbf{x}_i^t \vert\vert < r_i$
where $r_i$ is the maximal observation range of agent $i$.
Non-neighbor agents are ignored when we compose the local observation 
$\mathcal{O}_{i}^t$.
Note that we use the attention mechanism only for past observations $t\leq T$. In the case of the backward recurrent network used in Eq.~\ref{eq:backwards}, we simply set $w_{j \vert i}^t = 1$ for all neighbors to form $\mathcal{O}_{i}^t$ for $t>T$.

To represent the past observation sequence while embedding the target agent's navigation strategy, we employ
an 
RNN to encode the observations sequentially via the state variable $\mathbf{q}_i^t$, i.e. 
$\mathcal{O}_{i}^{1:t} := \mathbf{q}_i^t$, 
with $\mathbf{q}_i^t$ updated recurrently through
\begin{equation}
    \mathbf{q}_i^{t+1} = g(\mathcal{O}_{i}^{t+1}, \mathbf{q}_i^t).
\end{equation}
The initial state $ \mathbf{q}_i^1$ is extracted from the agent's and its neighbors' initial positions at $t=1$: 
\begin{equation}
    \mathbf{q}_i^1 = \sum\nolimits_j f_{\text{init}}(\mathbf{x}_j^1 - \mathbf{x}_i^1),
\end{equation}
where $f_{\text{init}}$ is a feature extraction neural network.

The attention weights, $w_{j \vert i}^t$, are obtained by a graph attention mechanism~\cite{velivckovic2017graph}, 
encoding node features by learnable edge weights.
To synthesize neighbors at time step $t$ based on observations from agent $i$, we regard agent $i$ and its neighbors as nodes in a graph with directed edges from the neighbor to the agent.
Attention weights corresponding to the edge weights are computed by
\begin{equation}\label{eq:att_weight}
    w_{j\vert i}^t = \frac{\exp(e_{j \vert i}^t)}{\sum_{k \neq i} \exp(e_{k \vert i}^t)}, 
\end{equation} 
where $e_{j \vert i}^t$
is the edge weight from the neighbor node $j$ to the agent node $i$.
To obtain the edge weights, following~\cite{sways2019}, we define the social features $\mathbf{k}_{j \vert i}^t$ of a neighbor $j$ observed by the agent $i$ at time step $t$ using three geometric features:
(1) the Euclidean distance between agents $i$ and $j$, (2) the cosine value of the bearing angle from agent $i$ to neighbor $j$, and (3) the minimal predicted distance~\cite{olivier2012minimal} from agent $i$ to $j$ within a given time horizon.

Given the neighbor's social features $\mathbf{k}_{j\vert i}^t$ and the agent's navigation features  $\mathbf{q}_i^{t-1}$, 
we compute the edge weight through the cosine similarity:
\begin{equation}
    e_{j\vert i}^t = \text{LeakyReLU}(f_{\mathbf{q}}(\mathbf{q}_{i}^{t-1}) \cdot f_\mathbf{k}(\mathbf{k}_{j\vert i}^t)),
\end{equation}
where $f_{\mathbf{q}}$ and $f_\mathbf{k}$ are neural networks. 
This leads to a dot-product attention using social features to synthesize neighbors while modeling the observations of an agent as a graph.
Here, 
$\mathbf{q}_i^{t-1}$, $\{\mathbf{k}_{j \vert i}^t\}$ and $\{f_{\mathbf{n}}(\mathbf{n}_{j \vert i}^t)\}$ correspond to the query, key and value vectors, respectively, in the vanilla attention mechanism
(cf. Fig~\ref{fig:overview}). 

In contrast to prior works~\cite{sgan2018,sadeghian2019sophie,sways2019} that apply attention mechanisms only on the last frame of the given observation sequence,
our approach computes attention at every time step. This allows to better extract an agent's navigation strategy while always taking into account the social influence from the neighbors. 
While prior works use RNN hidden states to represent each neighbor, 
our approach relies only on the neighbors' observed states, allowing to support sparse interactions where a neighbor is not consistently tracked.

\subsection{Final Position Clustering}
\begin{wrapfigure}[8]{r}{.45\linewidth}
\centering
    \raisebox{0pt}[\dimexpr\height-4\baselineskip\relax]{\includegraphics[width=.9\linewidth]{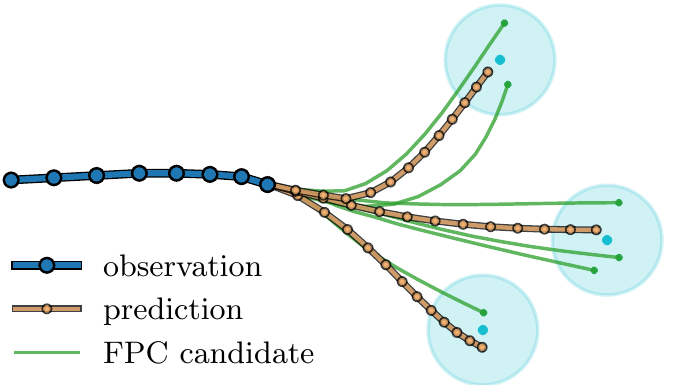}}
    \caption{An example of FPC to extract 3 predictions (orange) from 9 samples.}
    \label{fig:fpc_demo}
\end{wrapfigure}
The combination of the timewise VAE approach from~\ref{subsec:architecture} and the observation 
encoding from~\ref{subsec:observation_encoding} defines our vanilla SocialVAE model. It produces as many trajectory samples from the predictive distribution as required to infer the distribution over an agent's future position. 
However, when only a limited number of prediction samples are drawn from the distribution, 
bias issues may arise as some samples may fall into low-density regions or too many samples may be concentrated in high-density regions. 
As such, we propose \textit{Final Position Clustering} (FPC) as an optional postprocessing technique to improve the prediction diversity. 
With FPC, we sample at a higher rate than the desired number of predictions $K$.
Then we run a $K$-means clustering on the samples' final positions and for each of the $K$ clusters, we keep only the sample whose final position is the closest to the cluster's mean. This generates $K$ predictions in total as the final result.
Figure~\ref{fig:fpc_demo} shows an example of FPC selecting 3 predictions (orange) from 9 samples. Green trajectories depict discarded prediction samples 
and cyan regions are the clustering result from the final positions of these 9 samples.

\section{Experiments}\label{sec:exp}
In the following, we first give details on the implementation of SocialVAE and introduce the metrics used for evaluation. Then, we report and analyze experiments on 
two standard datasets: ETH/UCY~\cite{pellegrini2009you,lerner2007crowds} and  SDD~\cite{Robicquet2016}, and on the SportVU NBA
movement dataset~\cite{SportVU,Yue2014} which is a more challenging benchmark containing a rich set of complex human-human interactions. 

\noindent\textbf{Implementation details.} 
Our approach uses a local observation encoding and ignores agents' global coordinates.
To eliminate the global moving direction preferences,
we apply two types of data augmentation: flipping and rotation.
We employ GRUs~\cite{cho2014learning} as the RNN structure. 
The slope of the LeakyReLU activation layer is 0.2.
The time horizon used for the minimal predicted distance is 7s. 
The latent variables $\mathbf{z}_i^t$ are modeled as 32-dimensional, Gaussian-distributed variables.
All models are trained with 8-frame observations ($T=8$) and 12-frame predictions ($H=12$). 
The choice of sampling rate $K$ provides a tradeoff between evaluation performance and runtime and varies per dataset.  An upper bound is set at $50$.
We refer to our code repository
for our implementation, hyperparameters values and pre-trained models.
During inference, vanilla SocialVAE runs in real time on a machine equipped with a V100 GPU, 
performing 20 stochastic predictions within 0.05s for a batch of 1,024 observation samples.

\noindent\textbf{Evaluation Metrics and Baselines.} 
To evaluate our approach, we consider 
a deterministic prediction method, \textit{Linear}, which considers agents moving with constant velocities, and
the current state-of-the-art (SOTA) baselines of stochastic prediction in human-human interaction settings, as shown in Table~\ref{tab:results}.
Following the literature,  
we use best-of-20 predictions to compute \textit{Average Displacement Error}~(ADE) and \textit{Final Displacement Error}~(FDE) as the main metrics, and also report the \textit{Negative Log Likelihood}~(NLL) estimated from 2,000 samples. 
We refer to the appendix for details about these metrics. 
Among the considered baselines, PECNet~\cite{pecnet2020}, AgentFormer~\cite{yuan2021agentformer} and MemoNet~\cite{xu2022remember} rely on postprocessing to improve the model performance.
The reported numbers for 
Trajectron++~\cite{salzmann2020trajectron++}, BiTraP~\cite{yao2021bitrap} and SGNet-ED~\cite{wang2022stepwise} were obtained by training the models from scratch using the publicly released code 
after fixing a recently reported issue~\cite{Makansi2021,zamboni2022pedestrian,bae2022non} which leads to performance discrepancies compared to the numbers mentioned  in the original papers.

\subsection{Quantitative Evaluation}
\begin{table}[t]
\footnotesize
    \caption{Prediction errors reported as ADE/FDE in meters for ETH/UCY and in pixels for SDD.
    The reported values are the mean values of ADE/FDE using the best of 20 predictions for each trajectory.
    }
    \label{tab:results}
    \centering
    \setlength\tabcolsep{0.08cm}
    \begin{tabular}{l|ccccc|c}
         \toprule
         & \textbf{ETH} & \textbf{Hotel} & \textbf{Univ} & \textbf{Zara01} & \textbf{Zara02} & \textbf{SDD} \\
        \midrule
         Linear & 1.07/2.28 & 0.31/0.61 & 0.52/1.16 & 0.42/0.95 & 0.32/0.72 & 19.74/40.04 \\
         \hline
         SocialGAN & 0.64/1.09 & 0.46/0.98 & 0.56/1.18 & 0.33/0.67 & 0.31/0.64 & 27.23/41.44 \\
         SoPhie & 0.70/1.43 & 0.76/1.67 & 0.54/1.24 & 0.30/0.63 & 0.38/0.78 & 16.27/29.38 \\
         SocialWays & 0.39/0.64 & 0.39/0.66 & 0.55/1.31 & 0.44/0.64 & 0.51/0.92 & - \\
         \hline
         STAR & \textbf{0.36}/0.65 & 0.17/0.36 & 0.31/0.62 & 0.29/0.52 & 0.22/0.46 & -\\
         TransformerTF & 0.61/1.12 & 0.18/0.30 & 0.35/0.65 & 0.22/0.38 & 0.17/0.32 & -\\
         \hline
         MANTRA & 0.48/0.88 & 0.17/0.33 & 0.37/0.81 & 0.22/0.38 & 0.17/0.32 & \phantom{0}8.96/17.76 \\
         MemoNet$^\dagger$ & 0.40/0.61 & \textBF{0.11}/\textBF{0.17} & 0.24/0.43 & 0.18/0.32 & 0.14/0.24 & \phantom{0}8.56/12.66\\
         \hline
         PECNet$^\dagger$ & 0.54/0.87 & 0.18/0.24 & 0.35/0.60 & 0.22/0.39 & 0.17/0.30 & \phantom{0}9.29/15.93\\
         Trajectron++$^\ast$ & 0.54/0.94 & 0.16/0.28 & 0.28/0.55 & 0.21/0.42 & 0.16/0.32 & 10.00/17.15 \\
         AgentFormer$^\dagger$ & 0.45/0.75 & 0.14/0.22 & 0.25/0.45 & 0.18/0.30 & 0.14/0.24 & - \\
         BiTraP$^\ast$ & 0.56/0.98 & 0.17/0.28 & 0.25/0.47 & 0.23/0.45 & 0.16/0.33 & \phantom{0}9.09/16.31 \\
         SGNet-ED$^\ast$ & 0.47/0.77 & 0.21/0.44 & 0.33/0.62 & 0.18/0.32 & 0.15/0.28 & \phantom{0}9.69/17.01 \\
         \hline
        \midrule
         SocialVAE & 0.47/0.76 & 0.14/0.22 & 0.25/0.47 & 0.20/0.37 & 0.14/0.28 & \phantom{0}8.88/14.81 \\
          SocialVAE+FPC & 0.41/\textBF{0.58} & 0.13/0.19 & \textBF{0.21}/\textBF{0.36} & \textBF{0.17}/\textBF{0.29} & \textBF{0.13}/\textBF{0.22} & \phantom{0}\textBF{8.10}/\textBF{11.72}\\
         \bottomrule
    \end{tabular}
    \begin{tablenotes}
    \item $^\ast$: reproduced results with a known issue fixed
    \item $^\dagger$: baselines using postprocessing
    \end{tablenotes}
\end{table}
\noindent\textbf{Experiments on ETH/UCY.}
ETH/UCY benchmark~\cite{pellegrini2009you,lerner2007crowds} contains trajectories of 1,536 pedestrians recorded in five different scenes. 
We use the same preprocessing and evaluation methods as in prior works~\cite{sgan2018,sways2019}
and apply leave-one-out cross-validation.
Table~\ref{tab:results} shows the ADE/FDE results in meters.
As it can be seen, the linear model struggles to capture the complex movement patterns of pedestrians, with high errors
in most test cases,
though some of its results~\cite{SchollerALK20} are better than
the GAN-based approaches of SocialGAN~\cite{sgan2018}, SoPhie~\cite{sadeghian2019sophie} and SocialWays~\cite{sways2019}.
STAR~\cite{yu2020spatio} and TransformerTF~\cite{giuliari2021transformer} use Transformer architectures as a model backbone. 
While both Transformer-based approaches outperform the listed GAN-based approaches, with STAR achieving the best performance on ETH, they tend not to be better than the VAE-based models of Trajectron++, BiTraP and SGNet-ED.  
Compared to conditional-VAE baselines, 
our model leads to better performance, both with and without FPC.
Specifically, without FPC, the improvement of SocialVAE 
over SGNet-ED is about $12\%$ both on ADE and FDE. 
With FPC, SocialVAE improves SGNet-ED
by $21\%$ and $30\%$ in ADE and FDE, respectively. 
Compared to approaches that require postprocessing,
SocialVAE+FPC brings an improvement around $9\%$ on ADE and $13\%$ on FDE over AgentFormer, and $5\%$ on FDE over MemoNet 
while allowing for faster inference speeds and without requiring any extra space for memory storage.
The NLL evaluation results in Table~\ref{tab:nll} further show that the SocialVAE predictive distributions have superior quality, with higher probability (lower NLL) on the GT trajectories. For a fair comparison, we restrict here our analysis on SOTA methods from Table~\ref{tab:results} that do not rely on postprocessing. 

\noindent\textbf{Experiments on SDD.}
SDD~\cite{Robicquet2016} includes trajectories of 5,232 pedestrians in eight different scenes. 
We used the TrajNet~\cite{Becker2018} split to perform the training/testing processes and 
converted original pixel coordinates into spatial coordinates defined in meters for training. 
The related ADE/FDE and NLL results are reported in Tables~\ref{tab:results} and~\ref{tab:nll}, respectively. 
Following previous works, ADE and FDE are reported in pixels.
As shown in Table~\ref{tab:nll}, SocialVAE leads to more accurate trajectory distributions as compared to other VAE-based baselines. In addition, SocialVAE+FPC provides a significant improvement over existing baselines in terms of FDE as reported in Table~\ref{tab:results}.
Given that SDD has different homographies at each scene, to draw a fair comparison with results reported in meters, we refer to additional results in the appendix.

\begin{table}[t]
\footnotesize
    \centering
    \caption{NLL estimation on tested datasets.}
    \label{tab:nll}
    \setlength\tabcolsep{0.0475cm}
    \begin{tabular}{l|ccccc|c|cc}
        \toprule
         &  \textbf{ETH} & \textbf{Hotel} & \textbf{Univ} & \textbf{Zara01} & \textbf{Zara02} & \textbf{SDD} & \textbf{Rebounding} & \textbf{Scoring} \\
        \midrule
       Trajectron++  & 2.26 & -0.52 & 0.32 & -0.05 & -1.00 & 1.76 &  2.41 & 2.69 \\
       BiTraP  & 3.68 & 0.48 & 0.71 & 0.49 & -0.69 & 0.87 & 2.92 & 3.23 \\
       SGNet-ED & 2.65  & 0.92 & 1.36 & 0.13 & -0.77 & 1.53 & 3.28 & 3.05 \\
        \midrule
        \midrule
        SocialVAE & \textBF{0.96} & \textBF{-1.41} & \textBF{-0.49} & \textBF{-0.65} & \textBF{-2.67} & \textBF{-0.43} & \textBF{1.90} & \textBF{1.67} \\
        \bottomrule
    \end{tabular}
\end{table}

\noindent\textbf{Experiment on NBA Dataset.} 
We tested SocialVAE on the SportVU NBA movement dataset focusing on NBA games from the 2015-2016 season~\cite{SportVU,Yue2014}. Due to the large size of the original dataset, we extracted two sub-datasets to use as benchmarks named Rebounding and Scoring, consisting of 257,230 and 
\begin{wraptable}[10]{r}{.525\linewidth}
\raisebox{0pt}[\dimexpr\height-0.7\baselineskip\relax]{\begin{minipage}{\linewidth}
\footnotesize
    \centering
    \caption{ADE/FDE on NBA Datasets.}
    \label{tab:nba}
    \setlength\tabcolsep{0.1cm}
    \begin{tabular}{l|cc}
        \toprule
        Unit: feet & \textbf{Rebounding} &  \textbf{Scoring} \\
        \midrule
       Linear & 2.14/5.09 & 2.07/4.81 \\
       Trajectron++  & 0.98/1.93 & 0.73/1.46 \\
       BiTraP & 0.83/1.72 & 0.74/1.49 \\
       SGNet-ED &  0.78/1.55 &  0.68/1.30 \\
        \midrule
        \midrule
        SocialVAE & 0.72/1.37 & 0.64/1.17\\
        SocialVAE+FPC & \textBF{0.66}/\textBF{1.10} & \textBF{0.58}/\textBF{0.95} \\
        \bottomrule
    \end{tabular}
\end{minipage}}
\end{wraptable}
2,958,480 20-frame trajectories, respectively. We refer to the appendix for details on data acquisition. 
The extracted trajectories capture a rich set of agent-agent interactions and highly non-linear motions. 
Note that the overall frequency and the adversarial and cooperative nature of the interactions are significantly different from those in ETH/UCY and SDD, which makes trajectory prediction much more challenging~\cite{Makansi2021}. This is confirmed by the rather poor performance of the Linear baseline in such scenes (Table~\ref{tab:nba}).
SocialVAE achieves low ADE/FDE on both datasets, much better than the ones reported in prior work~\cite{Makansi2021,xu2022remember}, though it is unclear what training/testing data such works have used.   
Hence, we report our own comparisons to other VAE baselines in Tables~\ref{tab:nll} and~\ref{tab:nba}.  
Similar to ETH/UCY and SDD, SocialVAE exhibits SOTA performance on the two NBA datasets.

\subsection{Qualitative Evaluation}

\begin{figure*}[t]
    \centering
    \hfill\includegraphics[width=\linewidth]{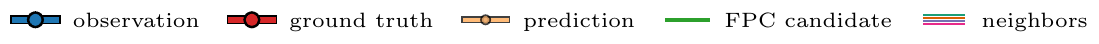}\\

    \includegraphics[width=.243\linewidth]{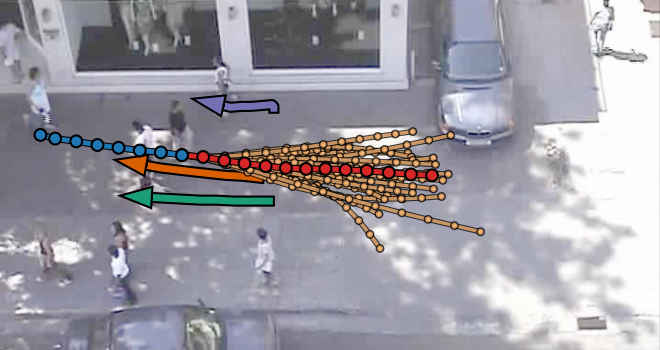}
    \hfill
    \includegraphics[width=.243\linewidth]{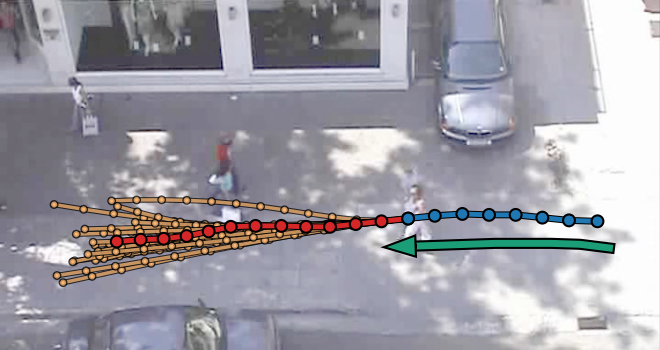}
    \hfill
    \includegraphics[width=.243\linewidth]{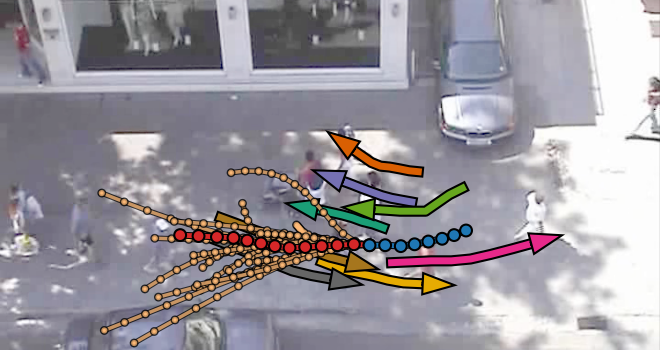}
    \hfill
    \includegraphics[width=.243\linewidth]{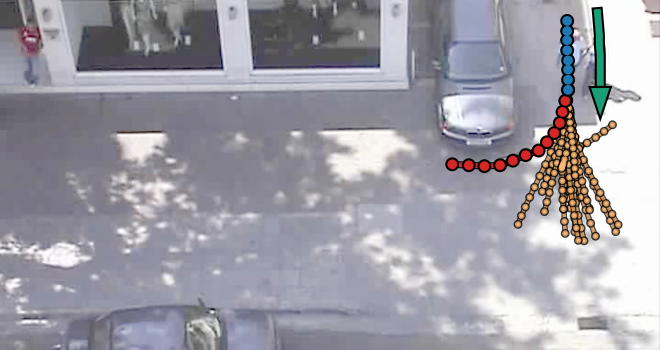}\\\vspace{0.0066\linewidth}
    
    \includegraphics[width=.243\linewidth]{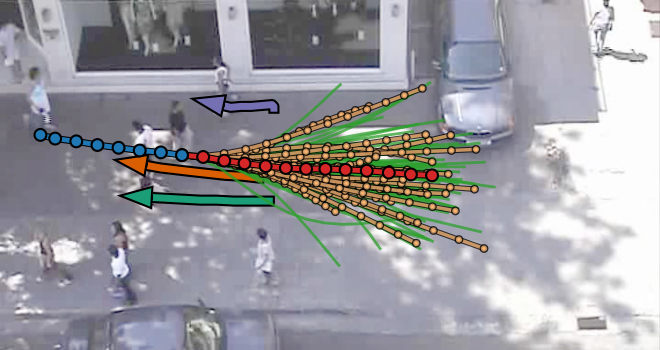}
    \hfill
    \includegraphics[width=.243\linewidth]{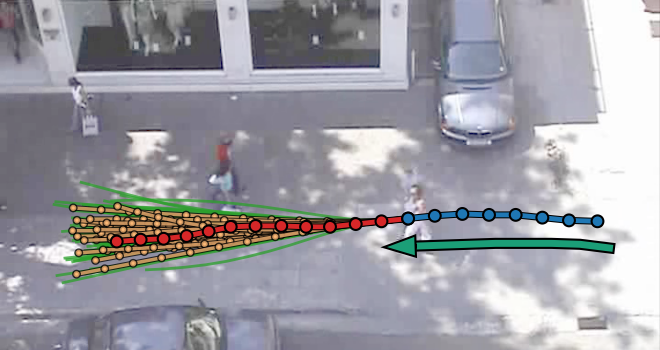}
    \hfill
    \includegraphics[width=.243\linewidth]{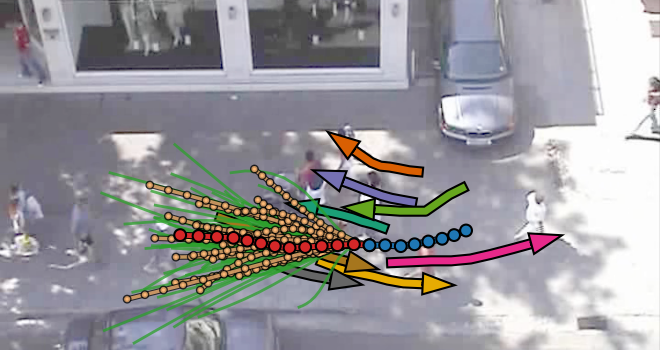}
    \hfill
    \includegraphics[width=.243\linewidth]{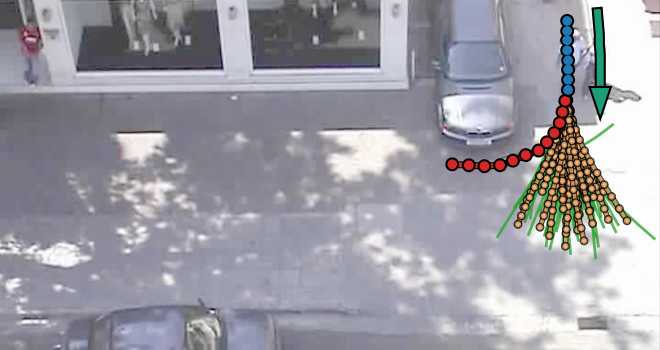}\\\vspace{0.0066\linewidth}
    
    \includegraphics[width=.243\linewidth]{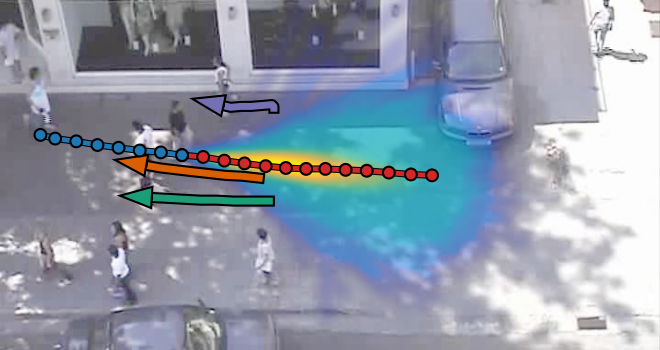}
    \hfill
    \includegraphics[width=.243\linewidth]{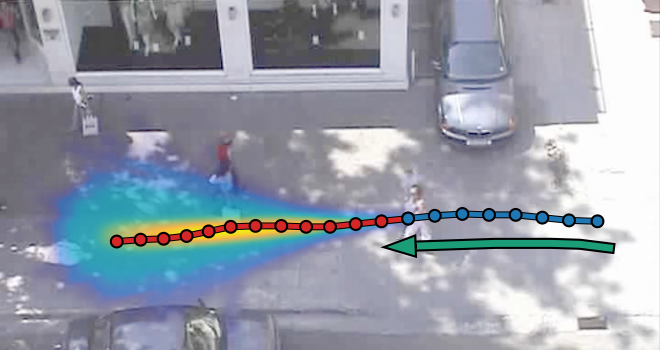}
    \hfill
    \includegraphics[width=.243\linewidth]{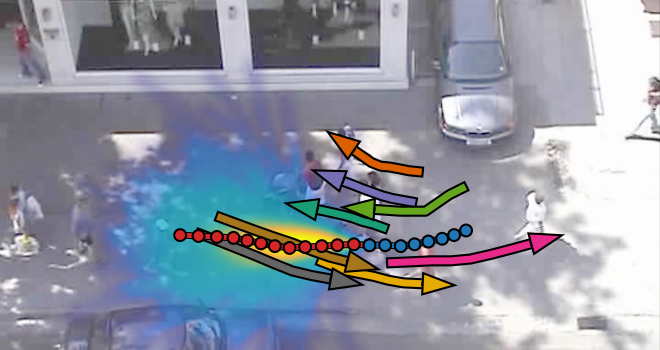}
    \hfill
    \includegraphics[width=.243\linewidth]{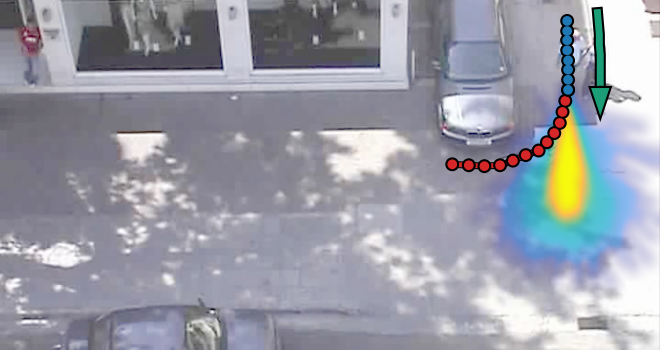}
    \caption{Examples of predictions from SocialVAE in the UCY \textit{Zara} scene. 
    Observed trajectories appear in blue, predicted trajectories appear in orange, and GT is shown in red.
    From top to bottom: 
    SocialVAE without FPC, SocialVAE+FPC, 
    and heatmaps of the predicted trajectory distribution. 
    Heatmaps are generated using 2,000 samples. 
    }
    \label{fig:example}
\end{figure*}
\begin{figure*}[t]
    \centering
    \includegraphics[width=.24\linewidth,trim={0 0.5cm 0 0.3cm},clip]{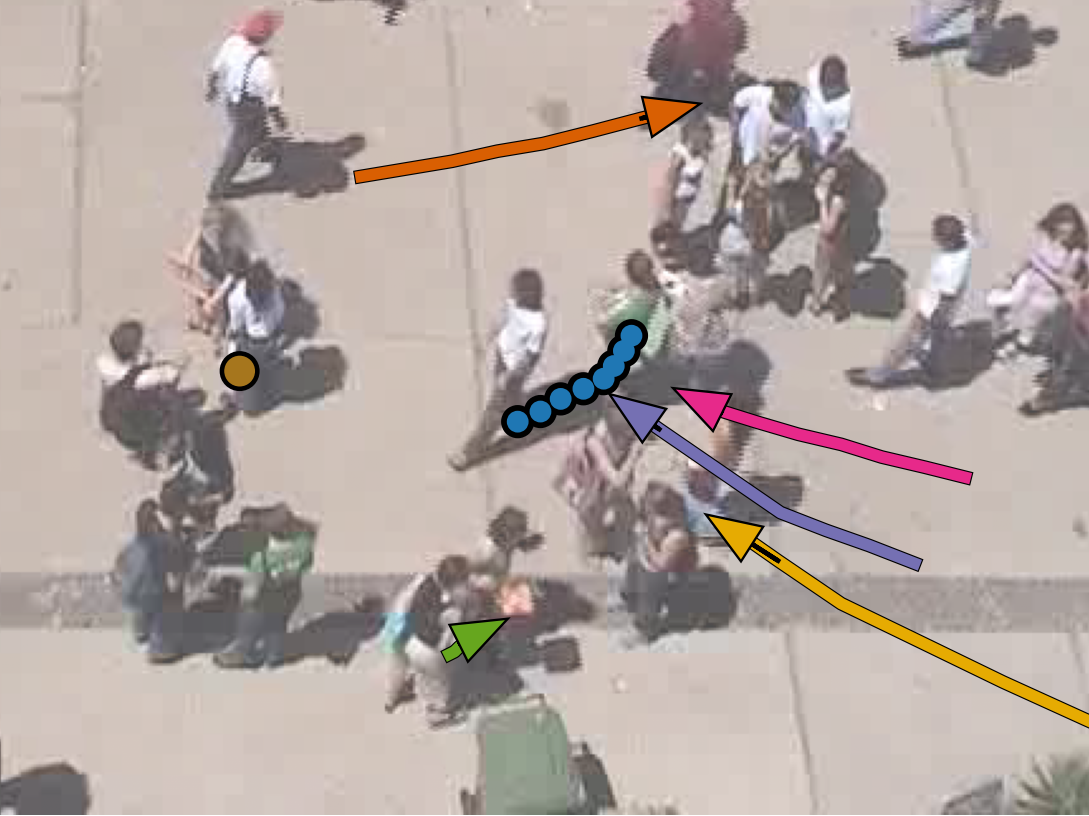}
    \includegraphics[width=.24\linewidth,trim={0 0.5cm 0 0.3cm},clip]{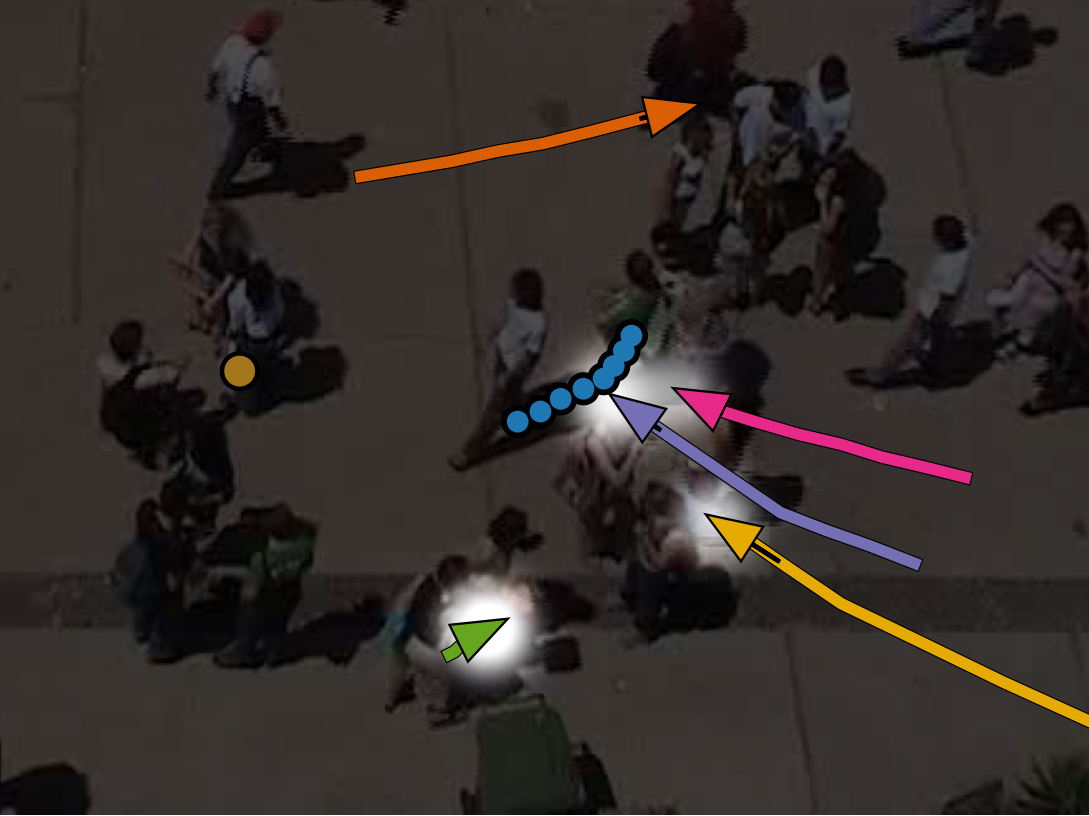}
    \hfill
    \includegraphics[width=.24\linewidth,trim={0 0.5cm 0 0.3cm},clip]{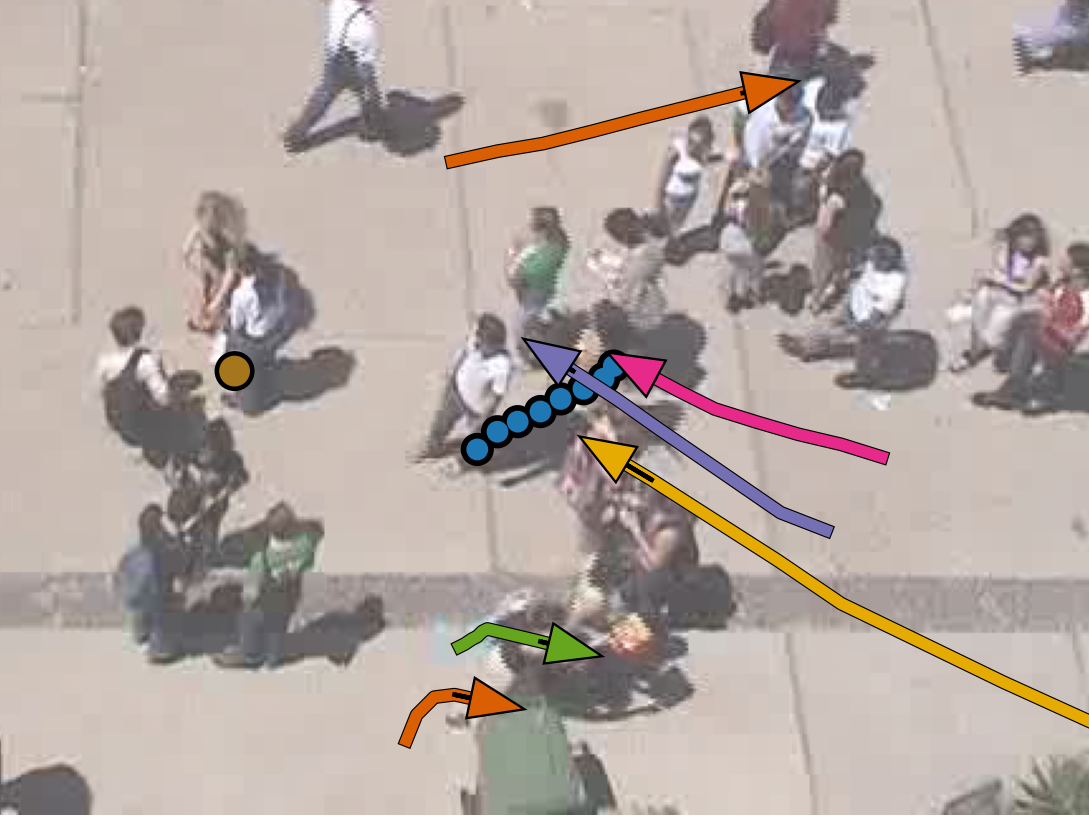}
    \includegraphics[width=.24\linewidth,trim={0 0.5cm 0 0.3cm},clip]{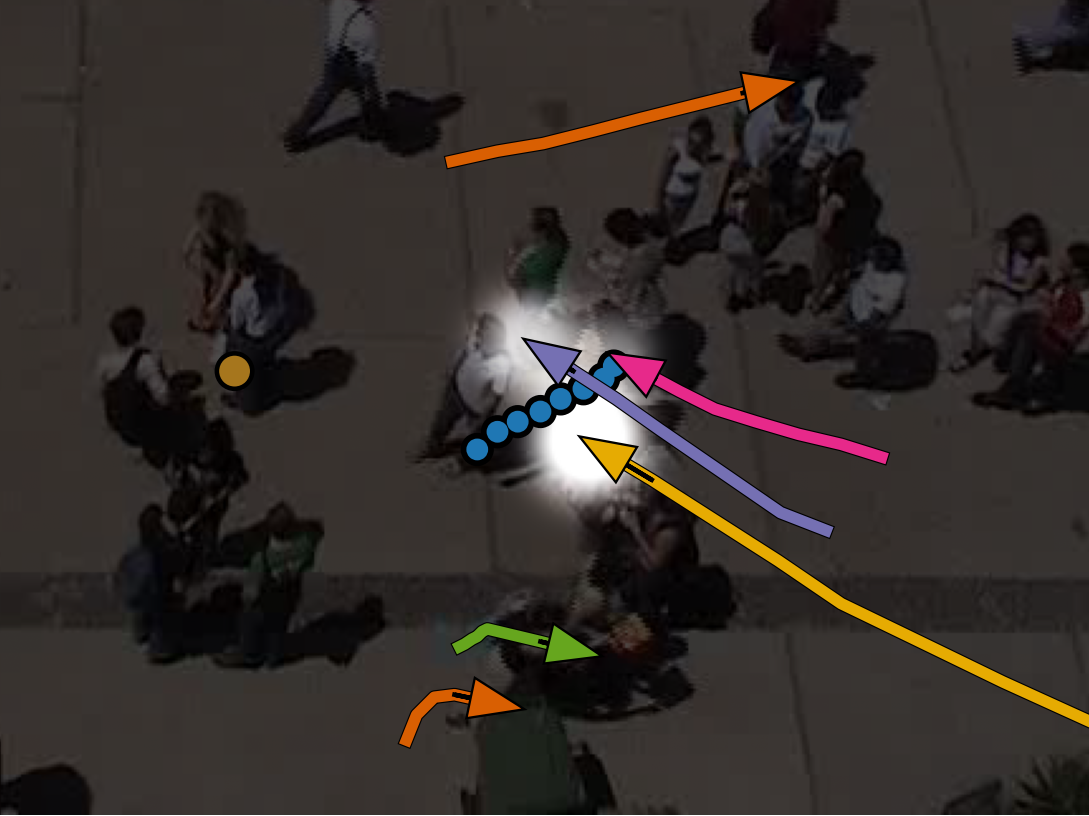}
    \caption{Attention maps at the 1st (left) and 20th (right) frames in UCY \textit{students003}. Trajectories of the pedestrian under prediction appear in blue. The other colored lines with arrows
    are the observed neighbors.
    Yellow dots denote stationary neighbors.
    }
    \label{fig:att}
\end{figure*}
\noindent{\textbf{Case Study on ETH/UCY.}}
Figure~\ref{fig:example} compares trajectories generated by SocialVAE with and without FPC in the Zara scene.
We show heatmaps of the predictive distributions in the 3rd row. They cover the GT trajectories very well in the first three scenarios.
In the 4th scenario, the agent 
takes a right turn that can be hardly captured from the 8-frame observation in which he keeps walking on a straight
line.
Though the GT trajectory is rather far from the average prediction,
SocialVAE's output distribution still partially covers it.
Using FPC improves the predictions diversity and helps to eliminate outliers.
For example, the topmost predictions in the 2nd and 3rd columns and the rightmost one in the 4th column are drawn from low-probability regions and eliminated by FPC.

To better understand how the social-feature attention mechanism impacts the prediction model,
we show the attention maps of scenes from \textit{students003} in Fig.~\ref{fig:att}. 
The maps are generated by visualizing the attention weights (Eq.\ref{eq:att_weight}) with respect to each neighbor as a white circle drawn on the location of that neighbor. 
The opacity and the radius of the circle 
increase 
with the weight of its associated neighbor.
As it can be seen, in the 1st frame, while monitoring the three neighbors on the right, much attention is paid to the green neighbor at the bottom who seems to be headed toward the pedestrian. 
In the 20th frame, the model ignores the green agent as it has changed its direction, and shifts its attention to the three nearby neighbors (red, purple, and yellow).
Among these neighbors, more attention is paid to the yellow agent walking toward the target pedestrian 
and less to the ones behind. 
In both scenes, the top-left, faraway neighbor is ignored along with the idle neighbors (yellow dots). 

\begin{figure*}[t]
    \centering
    \includegraphics[width=0.9\linewidth]{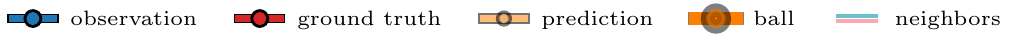}\hfill
    
    \includegraphics[width=.28\linewidth]{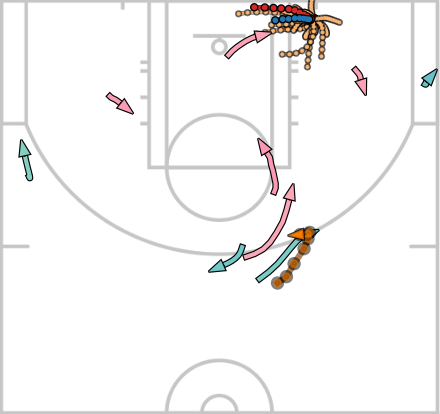}
    \includegraphics[width=.28\linewidth]{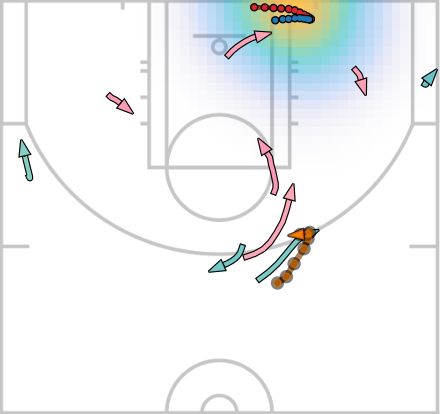}
    \includegraphics[width=.28\linewidth]{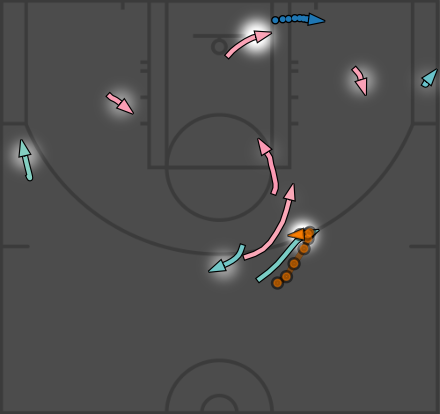}\\\vspace{0.0066\linewidth}
    
    \includegraphics[width=.28\linewidth]{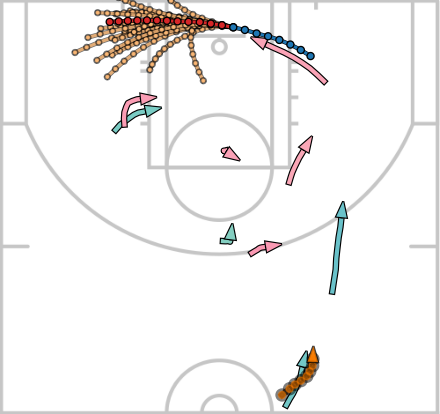}
    \includegraphics[width=.28\linewidth]{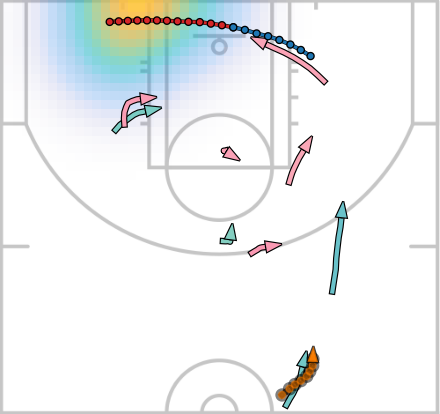}
    \includegraphics[width=.28\linewidth]{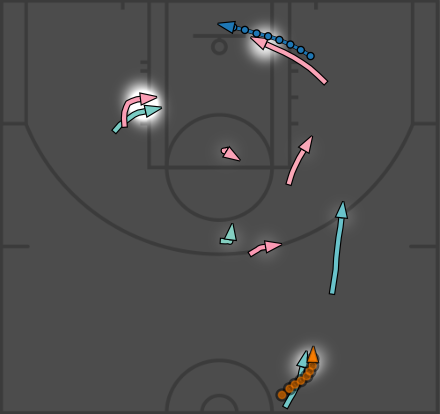}\\\vspace{0.0066\linewidth}
    
    \includegraphics[width=.28\linewidth]{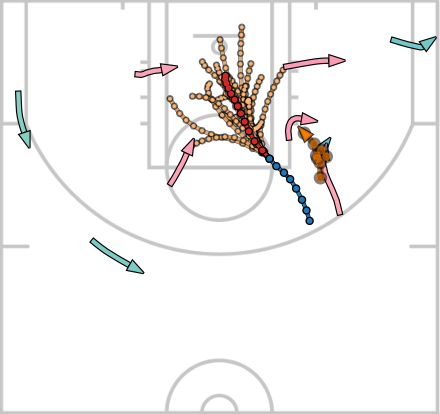}
    \includegraphics[width=.28\linewidth]{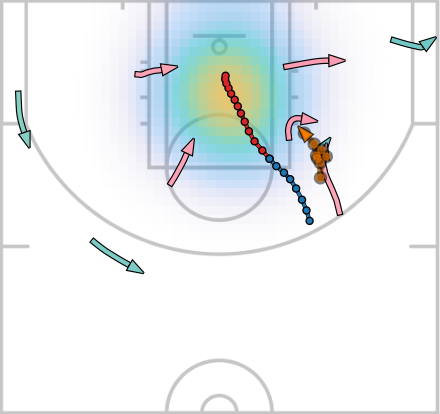}
    \includegraphics[width=.28\linewidth]{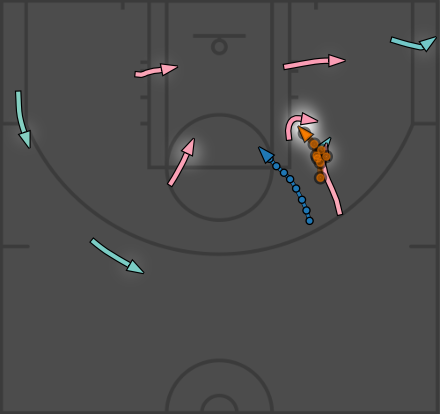}
    \caption{Examples of predictions from SocialVAE on NBA datasets. From left to right: predicted trajectories, distribution heatmaps and attention maps. 
    The azure lines with arrows
    denote trajectories of players who are in the same team as the one under prediction. 
    The pink ones denote opposing players.
    The orange indicates the ball.
    }
    \label{fig:nba_example}
\end{figure*}
\noindent{\textbf{Case Study on NBA Datasets.}}
The NBA scenarios 
contain rich interactions between players in both close and remote ranges. 
Players show distinct movement patterns with fast changes in heading directions.
Despite such complex behaviors, SocialVAE provides high-quality predictions, with distributions covering the GT trajectories closely.
In the challenging example shown in the 1st row of Fig.~\ref{fig:nba_example}, 
our model successfully predicts the player's intention to change his moving direction sharply and to shake his marker and catch the ball.
The predictive distribution gives several directions consistent with this intention and covers well the one that the player actually took.
From the attention map, we see that the model pays more attention to the defensive player who follows the one under prediction, and, across the range from the baseline to the three-point line, to the ball. 
Though we do not have any semantic information about the ball, 
our observation encoding approach helps the model identify its importance based on historical observations. 
Similar behaviors are found in the examples of the 2nd and 3rd rows, where the player performs a fast move to create a passing lane and to crash the offensive board, respectively. 
Note that our model pays more attention to teammates and opponents who influence the predicted agent's decision making. 
Furthermore, the predicted trajectories clearly exhibit multimodality, reflecting multiple responding behaviors that a player could take given the same scenario.

\subsection{Ablation Study}
\begin{table}[t]
\footnotesize
    \centering
    \setlength\tabcolsep{0.1cm}
    \caption{Ablation studies. ETH/UCY results are the average over five tested scenes.}
    \label{tab:ablation}
    \begin{tabular}{c|cccc|cccc}
       \toprule
       {$\uparrow^\text{a}$} (\%) & TL$^b$ & BP$^c$ & ATT$^d$ & FPC & \textbf{ETH/UCY} & \textbf{SDD}  & \textbf{Rebounding} & \textbf{Scoring}\\
       \midrule
        &  - & - & - & - & 
        0.35/0.50 & 13.43/17.81 & 1.02/1.48  & 1.03/1.43 \\
       14/3\phantom{0} & \ding{51} & - & - & - & 
        0.32/0.48 & 11.08/17.50 & 0.81/1.46 & 0.84/1.40\\
       20/6\phantom{0} &  - & \ding{51} & - & - & 
        0.30/0.47 & \phantom{0}9.45/16.01 & 0.78/1.46 & 0.74/1.39 \\
       26/11 &  \ding{51} & \ding{51} & - & - & 
        0.26/0.44 & \phantom{0}9.31/15.09 & 0.76/1.43 & 0.70/1.32 \\
       33/16 & \ding{51} & \ding{51} & \ding{51}  &  - & 
        0.24/0.42 & \phantom{0}8.88/14.81 & 0.72/1.37 & 0.64/1.17 \\
        38/32 & \ding{51} & \ding{51} & \ding{51} & \ding{51} & 0.21/0.33 & \phantom{0}8.10/11.72 & 0.66/1.10 & 0.58/0.95 \\
       \bottomrule
    \end{tabular}
    \begin{tablenotes}
    \item $^\text{a} \uparrow$: average performance improvement related the top row as baseline;
    \item $^\text{b}$ TL: timewise latents; $^\text{c}$ BP: backward posterior; $^\text{d}$ ATT: neighborhood attention.
    \end{tablenotes}
\end{table}
SocialVAE has four key components: timewisely generated latent variables (TL), backward posterior approximation (BP), neighborhood attention using social features (ATT), and an optional final position clustering (FPC).
To understand the contributions of these components, we present the results of ablation studies in Table~\ref{tab:ablation}.
For reference, the first row of the table shows the base performance of SocialVAE without using any of the key components.
Using either the TL scheme or the BP formulation reduces the ADE/FDE values, with the combination of the two leading to an average improvement of 26\%/11\% on ADE/FDE, respectively. 
Adding the ATT mechanism can further bring the error down.
In the last row, we also  report the performance when FPC is applied, highlighting the value of prediction diversity. As it can be seen, using all of the four components leads to a considerable decrease in FDE and SOTA performance (cf. Tables~\ref{tab:results} and~\ref{tab:nba}).
We refer to the supplementary material for an analysis on the FPC's sampling rate and explanatory visualizations of the latent space.

\section{Conclusion and Future Work}
We introduce SocialVAE as a novel approach for human trajectory prediction. 
It uses an attention-based mechanism to extract human navigation strategies from the social features exhibited in short-term observations,
and relies on a timewise VAE architecture using RNN structures to generate stochastic predictions for future trajectories. Our backward RNN structure in posterior approximation helps
synthesize 
whole trajectories for navigation feature extraction.
We also introduce FPC, a clustering method applied on the predicted trajectories final positions, 
to improve the quality of our prediction with a limited number of prediction samples.
Our approach shows state-of-the-art performance in most of the test cases from the ETH/UCY and SDD  
trajectory prediction benchmarks. We also highlighted the applicability of SocialVAE to SportVU NBA data. 
To further improve the prediction quality and generate physically acceptable trajectories, an avenue for future work is to introduce semantic scene information 
as a part of the model input. By doing so, our model could explicitly take both  human-space and 
human-agent interactions into account for prediction. 
This would also allow us to further evaluate SocialVAE on heterogeneous datasets~\cite{nuscenes,traphic}.
\\

\noindent{\textbf{Acknowledgements. }} This work was supported by the National Science Foundation under Grant No. IIS-2047632. 


\clearpage
%
%
\bibliographystyle{splncs04}
\bibliography{egbib}

\clearpage
\appendix

\section{Evaluation Metrics}
Below are the computation details of the evaluation metrics: 
\begin{itemize}
    \item \textit{Average Displacement Error}~(ADE), the Euclidean distance between a prediction trajectory $\{\mathbf{x}_i^t\}$ and the GT value $\{\mathbf{\hat{x}}_i^t\}$ averaged over all prediction frames for $t=T+1, \cdots, T+H$:
    \begin{equation}
        \text{ADE}(\{\mathbf{x_i}^t\}, \{\mathbf{\hat{x}}_i^t\}) = \frac{1}{H}\sum_{t=T+1}^{T+H} \vert\vert \mathbf{x}_i^t - \mathbf{\hat{x}}_i^t \vert\vert.
    \end{equation}
    
    \item \textit{Final Displacement Error}~(FDE), the Euclidean distance between the predicted position in the final frame and the corresponding GT value:
    \begin{equation}
        \text{FDE}(\{\mathbf{x_i}^t\}, \{\mathbf{\hat{x}}_i^t\}) = \vert\vert \mathbf{x}_i^{T+H} - \mathbf{\hat{x}}_i^{T+H}  \vert\vert.
    \end{equation}
    
    \item \textit{Negative Log Likelihood}~(NLL), the negative logarithm of the value of the predictive PDF at GT trajectories. The predictive distribution is obtained by Gaussian kernel density estimation from 2,000 samples. For simplicity, distributions at each time step are estimated independently and we use the joint distributions to compute PDF values.
\end{itemize}

\section{Social Features}
\begin{wrapfigure}[10]{r}{0.44\linewidth}
    \centering
    \vspace{-0.8cm}
    \includegraphics[width=\linewidth]{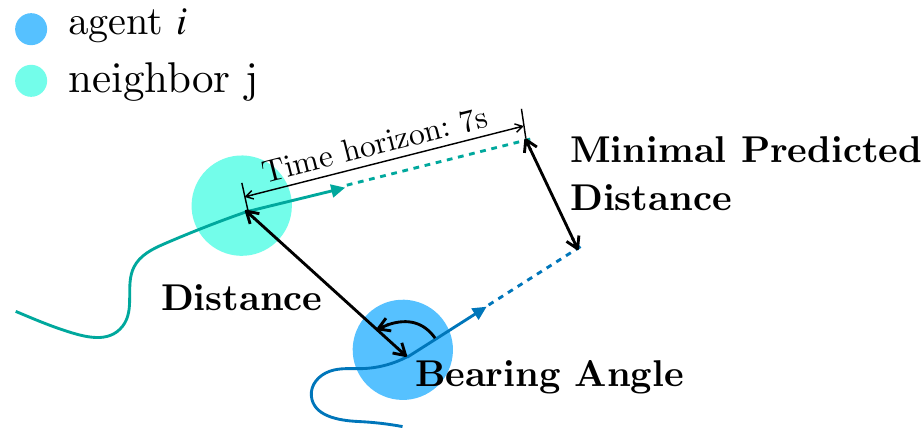}
    \caption{Demonstration of social features used for attention computation.}
    \label{fig:social_features}
\end{wrapfigure}
SocialVAE employs three social features for attention computation as shown in Fig.~\ref{fig:social_features}. Given an agent $i$ at time step $t$ and its neighbor $j$, these features are: 
\begin{itemize}
    \item the Euclidean distance between agents $i$ and $j$, i.e.
        $\vert\vert \mathbf{p}_{ji}^t \vert\vert$ where $\mathbf{p}_{ji}^t = \mathbf{x}_j^t - \mathbf{x}_i^t$;
    \item the cosine value of the bearing angle from agent $i$ to neighbor $j$, i.e.
        $\cos(\mathbf{p}_{ji}^t, \mathbf{d}_i^t)$;
    \item the minimal predicted distance~\cite{olivier2012minimal} from agent $i$ to $j$ within a time horizon $h$ (7s by default), i.e., 
        $\vert\vert \mathbf{p}_{ji}^t + \text{min}(\tau,h) \mathbf{v}_{ji}^t\vert\vert$, 
        where 
            $\mathbf{v}_{ji}^t = (\mathbf{d}_j^t - \mathbf{d}_i^t)/\Delta t$,
            $\tau = - (\mathbf{p}_{ji}^t \cdot \mathbf{v}_{ji}^t)/\vert\vert \mathbf{v}_{ji}^t \vert\vert^2$, and
            $\Delta t$ is the sampling interval between two frames.
\end{itemize}

\section{Data Acquisition of SportVU NBA Dataset}
\begin{table}[h]
    \caption{Statistical information on SportVU NBA Datasets.}
    \label{tab:nba_stats}
    \centering
    \setlength\tabcolsep{0.175cm}
    \begin{tabular}{r||c|c}
        \toprule
        & \textbf{Scoring} & \textbf{Rebounding} \\
        \midrule
         \# of Training/Testing Scenes & 2,979/744 & 3,754/938\\
        Avg. Play Duration (s) & 11.82 & 2.94 \\
        \# of Trajectories (20-frame) & 2,958,480 & 257,230 \\
        Avg. Trajectory Length (m) & 4.55 & 3.87 \\
        \bottomrule
    \end{tabular}
\vspace{-0.3cm}
\end{table}

To test our approach on scenarios with complex and intensive 
human-human interactions, we have extracted two sub-datasets from the SportVU basketball movement dataset~\cite{Yue2014,Makansi2021}
focusing on games from the 2015-2016 NBA regular season: 
\begin{itemize}
    \item {\bf{Rebounding dataset}}.
    This dataset focuses on scenes involving a missed shot with players moving to grab the rebound. The dataset contains a number of interesting interactions, including players boxing out their opponents to allow a team member to grab a rebound, players moving toward the basket, and players starting to run on the other side of the half court for offensive or defensive purposes. 
    \item {\bf{Scoring dataset}}. 
    This dataset focuses on scenes involving a team scoring a basket.
    The resulting dataset contains a rich set of player-player interactions, both cooperative and adversarial, including highly non-linear player motions, set plays employed by different teams, and different offensive and defensive schemes.
\end{itemize}

We refer to Table~\ref{tab:nba_stats} for detailed characteristics of the two datasets. 
For each dataset, scenes are randomly split into testing and training sets using a 1:4 ratio. 
The original data were recorded at 25 FPS with a time interval of 0.04s between frames.
In consideration that basketball players move much faster than normal pedestrians, 
we downsample the data to the time interval of 0.12s (instead of 0.4s that we use on ETH/UCY and SDD benchmarks). 
We employ the same network structure that we have used for the ETH/UCY and SDD benchmarks, 
and do 12-frame predictions for players (excluding the ball) based on 8-frame observations.
This leads to training and testing trajectories having 20 frames, with the average length around $4$m, as reported in Table~\ref{tab:nba_stats}.
The neighborhood radius 
is set such that the whole arena is covered, which means that all the players and the ball are taken into account during observation encoding.

\section{Additional Results on SDD}
\vspace{-0.8cm}
\begin{table}[h]
\footnotesize
    \caption{ADE/FDE in meters on SDD. The reported numbers are the mean value of the best-of-20 predictions.}
    \label{tab:sdd_meter}
    \centering
    \setlength\tabcolsep{0.175cm}
    \begin{tabular}{c|ccc||cc}
        \toprule
         & Trajectron++ & BiTraP & SGNet-ED & SocialVAE & SocialVAE+FPC\\
        \midrule
        \textbf{SDD} & 0.34/0.58 & 0.32/0.57 & 0.33/0.58 & 0.30/0.50 & \textBF{0.27}/\textBF{0.39} \\
        \bottomrule
    \end{tabular}
\vspace{-1cm}
\end{table}

\section{Sensitivity Analysis on FPC}
\begin{figure}[h]
    \vspace{-0.8cm}
    \centering
    \includegraphics[width=\linewidth]{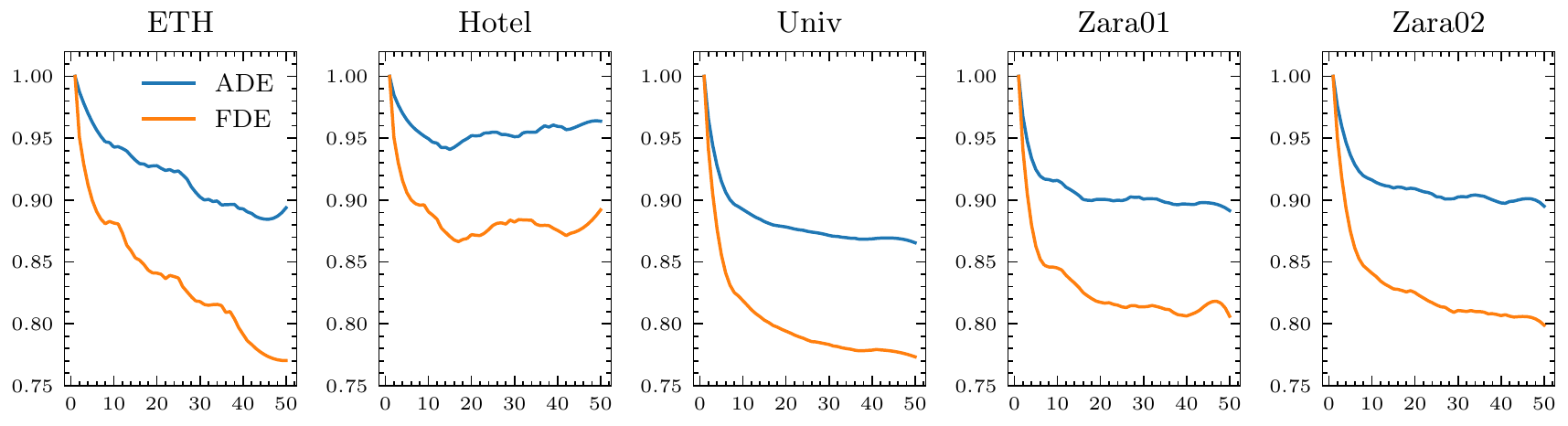}
    \caption{Performance of FPC with respect to different sampling rates (1-50). All values are normalized by that of sampling rate 1 (no FPC).
    }
    \label{fig:fpc}
\vspace{-0.5cm}
\end{figure}
\noindent Figure~\ref{fig:fpc} 
plots the ADE and FDE values when FPC is applied with varying sampling rate on the ETH/UCY benchmark.
As shown in the figure,
the errors decrease roughly as the sampling rate increases.
Typically, FPC can lead to a significant improvement about $10\%$ on ADE and $18\%$ on FDE within a sampling rate around 20.
Further increasing the sampling rate can only bring about a $2\%$  extra improvement (with the exception of a $5\%$ FDE improvement on ETH), at the cost though of higher running time.

\section{Latent Space Analysis}
\begin{figure}[h!]
    \vspace{-0.8cm}
    \centering
    \includegraphics[width=0.6\linewidth]{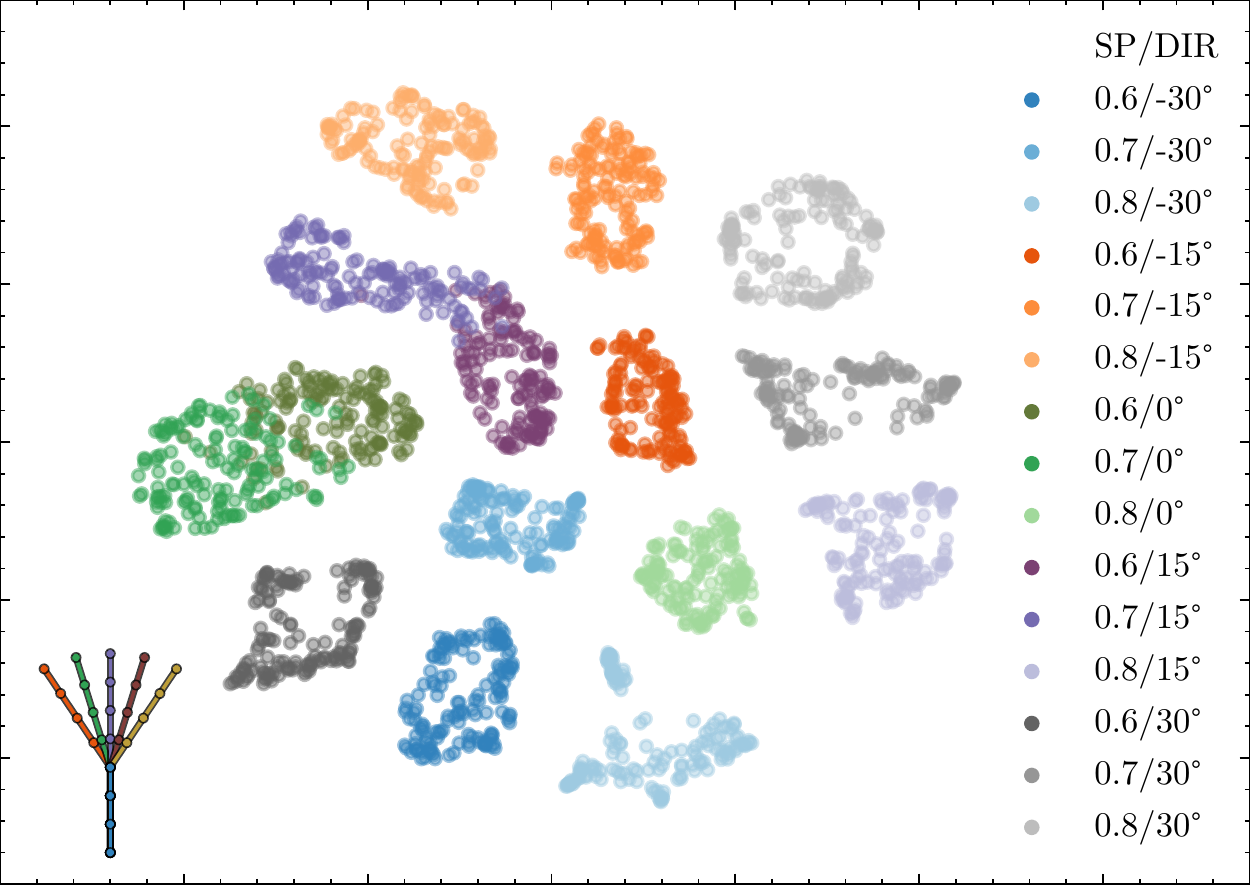}
    \caption{t-SNE visualization of latent variable distributions given varying observation trajectories with different speeds (SP) and turn directions (DIR). 
    The left bottom corner gives an example of the observed trajectories with different turn directions at the 5th frame from $-30^\circ$ to $30^\circ$. 
    For each of the five trajectory shapes, we consider observations with three constant speeds from $0.6$m to $0.8$m. This gives us a combination of 15 observations, as shown in the legend.}
    \label{fig:latent}
\vspace{-0.3cm}
\end{figure}
\noindent To show that our model can learn a structured embedding of the observed trajectories, we plot the latent variable distributions in
Fig.~\ref{fig:latent}.
To do so, we run a model pre-trained using the ETH/UCY datasets on 15 different 8-frame observations, which are the combinations of five distinct trajectory headings and three distinct speeds.
For each observation, we draw 150 samples of the latent variables from the prior at the first time step of prediction, i.e. $\mathbf{z}_i^{T+1}$. 
As it can be seen, our model can clearly distinguish observations with semantically different features.

\end{document}